\documentclass[graybox]{svmult}

\usepackage{mathptmx}       
\usepackage{helvet}         
\usepackage{courier}        
\usepackage{type1cm}        
\usepackage{makeidx}         
\usepackage{graphicx}        
\usepackage{multicol}        
\usepackage[bottom]{footmisc}

\usepackage{amsmath}
\usepackage{svg}
\usepackage[inline]{enumitem}
\usepackage{subfig}
\usepackage{amssymb}

\usepackage{caption}

\usepackage{color}
\usepackage{soul}
\sethlcolor{blue}

\newcommand{\lsmany}{DeepVO}
\newcommand{\lsmanyopt}{Nicolai \emph{et al.}}
\newcommand{\mcl}{Ground Truth}
\newcommand{\cnnBspp}{cnnBspp}
\newcommand{\mclgt}{Ours}

\newcommand{\secref}[1]{Section~\ref{#1}}

\renewcommand{\eqref}[1]{Equation~(\ref{#1})}
\newcommand{\figref}[1]{Figure~\ref{#1}}
\newcommand{\tabref}[1]{Table~\ref{#1}}

\newcommand{\boxplotlegend}{
	\scriptsize
	\begin{tabular}{ccccc}
	\raisebox{0.5ex}{\includegraphics[scale=1]{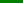}} \lsmany&
	\raisebox{0.5ex}{\includegraphics[scale=1]{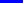}} \lsmanyopt&
	\raisebox{0.5ex}{\includegraphics[scale=1]{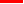}} \mcl&
	\raisebox{0.5ex}{\includegraphics[scale=1]{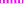}} 
\cnnBspp&
	\raisebox{0.5ex}{\includegraphics[scale=1]{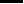}} \mclgt
	\end{tabular}
}

\newcommand{\GT}{Ground Truth}
\newcommand{\OursVO}{Ours-Metric}
\newcommand{\OursTM}{Ours-Topometric}

\newcommand{\boxplotlegendTopoMetric}{
	\scriptsize
	\begin{tabular}{ccc}
	\raisebox{0.5ex}{\includegraphics[scale=1]{legend_mcl.pdf}} \GT&
	\raisebox{0.5ex}{\includegraphics[scale=1]{legend_mclgt.pdf}} \OursVO&
	
\raisebox{0.5ex}{\includegraphics[scale=1]{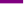}} \OursTM
	\end{tabular}
}

\newcommand{\Est}[1]{\hat{#1}}

\DeclareMathAlphabet{\mathcal}{OMS}{cmsy}{m}{n}

\def\argmin{\mathop{\rm argmin}}

\makeindex             

\begin{document}
\title*{Topometric Localization with Deep Learning}
\author{Gabriel L. Oliveira$^*$ \and Noha Radwan$^*$ \and Wolfram Burgard \and Thomas 
Brox}
\institute{$^*$These authors contributed equally. All authors are with the Department of Computer Science, University of 
Freiburg, Germany. Corresponding author's \email{oliveira@informatik.uni-freiburg.de}}
%
%
\maketitle

\newcommand{\abstracttext}[0]{
Compared to LiDAR-based localization methods, which provide high accuracy but rely on expensive sensors, visual localization 
approaches only require a camera and thus are more cost-effective while their accuracy and reliability typically is inferior to 
LiDAR-based methods. 
In this work, we propose a vision-based localization approach that learns from LiDAR-based localization methods by using their 
output as training data, thus combining a cheap, passive sensor with an accuracy that is on-par with LiDAR-based localization.
The approach consists of two deep networks trained on visual odometry and topological localization, respectively, and a 
successive 
optimization to combine the predictions of these two networks. 
We evaluate the approach on a new challenging pedestrian-based dataset captured over the course of six months in varying weather 
conditions with a high degree of noise. The experiments demonstrate that the localization errors are up to $10$ times smaller 
than 
with traditional vision-based localization methods.   
}

\abstract*{\abstracttext}
\abstract{\abstracttext}

\section{Introduction}

Robot localization is essential for navigation, planning, and autonomous operation. There are vision-based approaches addressing 
the robot localization and mapping problem in various environments~\cite{garcia2015vision}.
Constant changes in the environment, such as varying weather conditions and seasonal changes,
and the need of expert knowledge to properly lay out a set of domain specific features
makes it hard to develop a robust and generic solution for the problem. 

Visual localization can be distinguished according to two main types: 
\textit{metric} and \textit{topological} localization. Metric localization 
consists of computing the coordinates of the location of the observer. The 
coordinates of the vehicle pose are usually obtained by visual odometry 
methods~\cite{visapp15, melekhov2017cnnBspp,MohantyDeepVO,NicolaiRSSw2016}. 
Visual metric approaches can provide accurate localization values, but 
suffer from drift accumulation as the trajectory length increases. In general, they 
do not reach the same accuracy as LiDAR-based localization techniques.   
Topological localization detects the observer's approximate position from a 
finite set of possible locations~\cite{lowry2016visual, 
arandjelovic2016netvlad, arroyo2016fusion}. This class of localization methods 
provides coarse localization results, for instance, if a robot is in front 
of a specific building or room. Due to its limited state space, topological 
approaches provide reliable localization results without drift, but only
rough position measurements.    

\begin{figure}[t] 
	\centering
	\includegraphics[width=\textwidth]{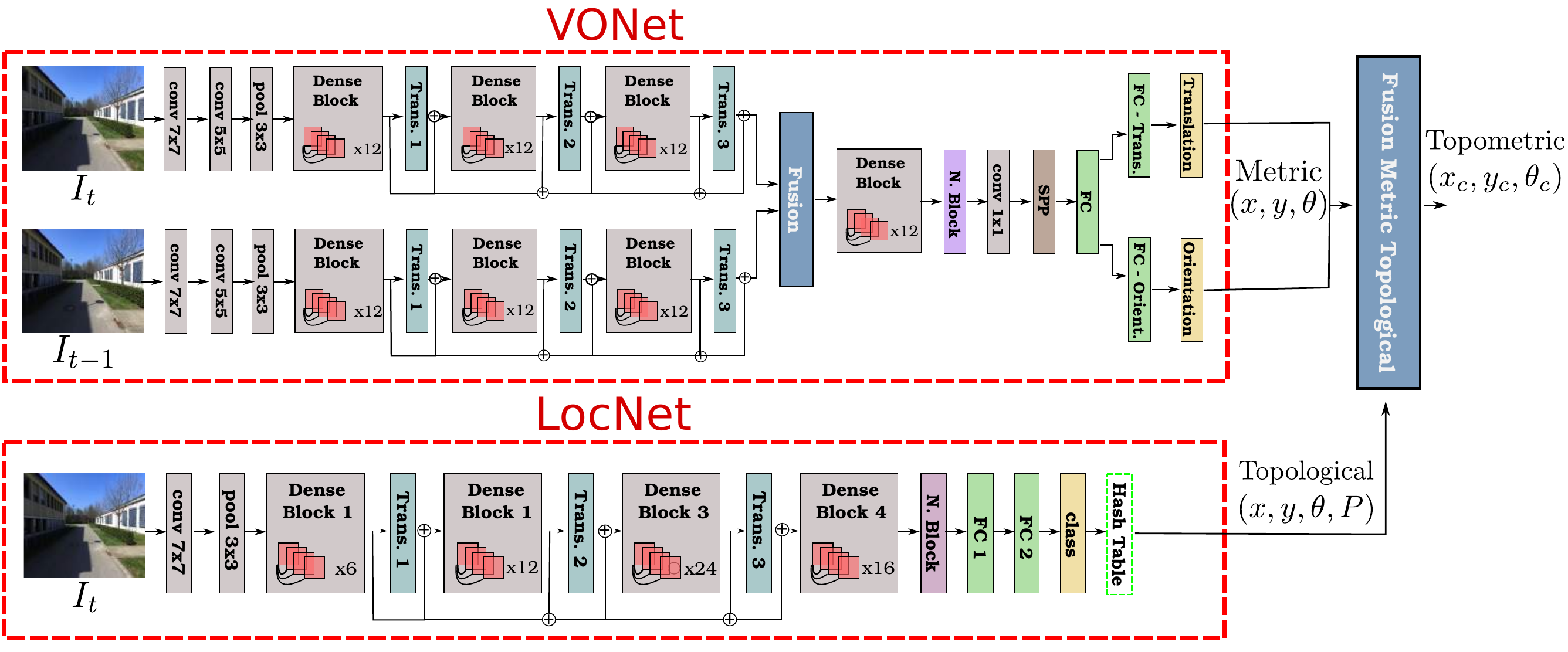}
	\caption{The three main modules of our approach:
          \textit{metric}, \textit{topological} and
          \textit{topometric}.  The metric part called VONet estimates
          odometry values given two consecutive images.  The network
          designed for metric localization is called VONet.  The
          topological module denoted as LocNet provides the closest
          topological location of an input image, mapping it back to a
          pose through a hash table. VONet and LocNet are trained
          independently of each other.  The optimization part fuses
          metric and topological predictions to produce the corrected
          topometric pose estimate.}
	\label{fig:SystemOverview}
\end{figure}

Accurate, drift-free localization can be obtained by combining both
approaches, which is known as \textit{topometric}
localization~\cite{mazuran15isrr}.  In this paper, we present a novel
topometric localization technique that formulates metric and
topological mapping as learning problems via a deep network.  
The first network computes the relative visual odometry between consecutive images, 
while the second network estimates the topological location. Both 
networks are trained independently and their outputs are provided to our 
topometric optimization technique. By fusing the output of the visual odometry 
network with the predicted topological location, we are able to produce an 
accurate estimate that is robust to the trajectory length and has minimum drift 
accumulation. An overview of the proposed method is shown in
\figref{fig:SystemOverview}. We introduce a real-world dataset collected from the Freiburg University Campus over the course of 
six month. We compare the proposed approach on the dataset, to state-of-the-art visual odometry methods. The experimental 
evaluation shows
that the proposed approach yields basically the same accuracy as LiDAR-based localization methods. The dataset will be made 
publicly available to simplify comparisons and foster research in visual localization.

\section{Related Work}

One of the seminal deep learning approaches for visual odometry was proposed by 
Konda  \emph{et al.}~\cite{visapp15}. They proposed a CNN architecture which 
infers odometry based on classification. A set of prior velocities and 
directions are classified through a softmax layer to infer the 
transformation between images from a stereo camera. A major drawback
of this approach lies in modeling a regression problem as a 
classification one which reduces the representational capabilities of the learned model.
Other approaches
\cite{melekhov2017cnnBspp,MohantyDeepVO,NicolaiRSSw2016} tackle the problem of 
visual ddometry as a regression problem. Nicolai  \emph{et al.}~\cite{NicolaiRSSw2016} proposed a CNN architecture 
for depth images based on LiDAR scans. They proposed a simple 
architecture that yields real-time capability. 
Mohanty  \emph{et al.}~\cite{MohantyDeepVO} proposed a Siamese AlexNet~\cite{AlexNet} 
based approach called DeepVO, where the translation and rotation outputs
of the network are regressed through an L2-loss layer with equal weight values.
Choosing weight values for the translational and rotational components of the
regression output is explored by Melekhov  \emph{et al.}~\cite{melekhov2017cnnBspp}. They propose a Siamese 
AlexNet~\cite{AlexNet} 
network similar to DeepVO~\cite{MohantyDeepVO}, where they
add a weight term to balance the translational and rotational errors. They
additionally use a spatial pyramid pooling (SPP) layer~\cite{SPP} which allows for arbitrary input image resolutions.
Our metric localization approach shares similarities with \cite{melekhov2017cnnBspp}, since we use SPP and a loss function which 
balances translation and rotation losses. Our contribution to this problem is a new densely connected architecture along with a 
different angular representation.


Another part of our approach is topological localization. With increasing focus on the long-term autonomy of mobile agents in 
challenging environments, the need for life-long
visual place recognition has become more crucial than before~\cite{lowry2016visual}.
In~\cite{arandjelovic2016netvlad}, the authors present an end-to-end approach for
large-scale visual place recognition. Their network aggregates mid-level convolutional features extracted from the entire image 
into a compact vector representation using VLAD \cite{jegou:VLAD}, resulting in a compact and robust image descriptor.
Chen \emph{et al.}~\cite{chen2014convolutional} combine a CNN with spatial
and sequential filtering. Using spatio-temporal filtering and spatial continuity checks ensures
consecutive first ranked hypotheses to occur in close indices to the query image.
In the context of this work, the problem of visual place recognition can be 
considered as a generalized form of topological localization. 
Similar to the visual place recognition approaches presented above, the authors in
\cite{arroyo2016fusion} present a topological localization approach
that is robust to seasonal changes using a CNN. They fuse information
from convolutional layers at several depths, finally compressing the output into a single feature vector. Image matching is done 
by computing the Hamming distance between the feature vectors after binarization, thus improving the speed of the whole approach.
Similar to these approaches, we use a CNN architecture that aggregates information from convolutional layers to learn a compact 
feature representation. However, instead of using some distance heuristic, the output of our network is a probability 
distribution 
over a discretized set of locations. Moreover, whereas previous methods rely on either distance or visual
similarity to create the topological set of locations, we introduce a technique that
takes both factors into account.


Topometric localization was explored by the works of Badino  \emph{et al.} \cite{Badino11Topometric} and Mazuran  \emph{et al.} 
\cite{mazuran15isrr}. Badino 
 \emph{et al.} \cite{Badino11Topometric} proposed one of the first methods to 
topometric localization. They represent visual features 
using a topometric map and localize them using a discrete Bayes filter. The topometric 
representation is a topological map where each node is linked to a pose in the 
environment. Mazuran  \emph{et al.} \cite{mazuran15isrr} extended the previous 
method to relative topometric localization. They introduced a topometric 
approach which does not assume the graph to be the result of an optimization 
algorithm and relaxed the assumption of a globally consistent metric map.



While Mazuran \emph{et al.}~\cite{mazuran15isrr} use LiDAR measurements and Badino  \emph{et al.}~\cite{Badino11Topometric} rely 
on a multi-sensory approach, which employs camera, GPS, and an inertial sensor unit, our approach is based on two CNNs for metric 
and topological estimation and only requires visual data. 

%

\section{Methodology}

This paper proposes a topometric localization method using image sequences
from a camera with a deep learning approach. The topometric localization 
problem consists of estimating the robot pose  $\mathbf{x}_t \in 
SE(2)$ and its topological node $\mathbf{n}_t \in 
SE(2)$, given a map encoded as a set of globally referenced nodes equipped with 
sensor readings~\cite{mazuran15isrr,sprunk12icra}.

We propose a deep CNN to estimate the relative motion between two consecutive 
image sequences, which is accumulated across the traversed path to provide a 
visual odometry solution. In order to reduce the drift often encountered by 
visual odometry, we propose a second deep CNN for visual place recognition. The 
outputs of the two networks are fused to an accurate location estimate.
In the remainder of this section, we describe the two networks and the fusion approach in detail. 
\subsection{Metric Localization}
\label{sec:metricMeth}

The goal of our metric localization system is to estimate the relative camera 
pose from images. We design a novel architecture using
\emph{Dense-blocks}~\cite{huang2016densely} as base, which given a pair of images in a 
sequence $(I_{t},I_{t-1})$ will predict a \textit{4-dimensional} relative camera pose 
$\mathbf{p}_t$:

\[
\mathbf{p}_t := [\Delta{\mathbf{x}_t^{tr}}, \Delta{\mathbf{r}_t}]
\]

where $\Delta{\mathbf{x}_t^{tr}}:=\mathbf{x}^{tr}_{t}-\mathbf{x}^{tr}_{t-1} 
\in \mathbb{R}^2$ is the $x$ and $y$ relative translation values and 
$\Delta{\mathbf{r}_t}:=[\sin(\mathbf{x}_t^{\theta}-\mathbf{x}_{t-1}^{\theta}), 
\cos(\mathbf{x}_t^{\theta}-\mathbf{x}_{t-1}^{\theta})] \in \mathbb{R}^2$
 is the relative rotation estimation.
We represent the rotation using Euler6 notation, i.e., the rotation
angle $\theta$ is represented by two components $[\sin(\theta)$,$\cos(\theta)]$. 
 
 

\subsubsection{Loss Function}
\label{lossfunction}
The proposed metric localization network is designed to regress the relative 
translation and orientation of an input set $(I_t,I_{t-1})$. We train the network based on 
the Euclidean loss between the estimated vectors and the ground truth. 
Having a loss function that deals with both the translation and orientation in the same 
manner was found inadequate, due to the difference in the scale between them. Instead, we define the following loss function:

\begin{equation}
  \mathcal{L} := \left \| \Delta\Est{\mathbf{x}}^{tr} - \Delta \mathbf{x}^{tr}  
\right 
\|_2 + \beta \left 
\| \Delta\Est{\mathbf{r}} - \Delta \mathbf{r}  \right \|_2,
\end{equation}
where  $\Delta \mathbf{x}^{tr}$ and $ \Delta \mathbf{r}$ are respectively the 
relative ground-truth translation and rotation vectors and $\Delta\Est{\mathbf{x}}^{tr}$ and 
$\Delta\Est{\mathbf{r}}$ their 
estimated counterparts. Similar to Kendall  \emph{et al.}~\cite{kendall2015posenet}, we use the parameter $\beta>0$ to balance 
the 
loss for the translation and orientation error. 

\subsubsection{Network Architecture}

\begin{figure}[t] 
	\centering
	\includegraphics[width=\textwidth]{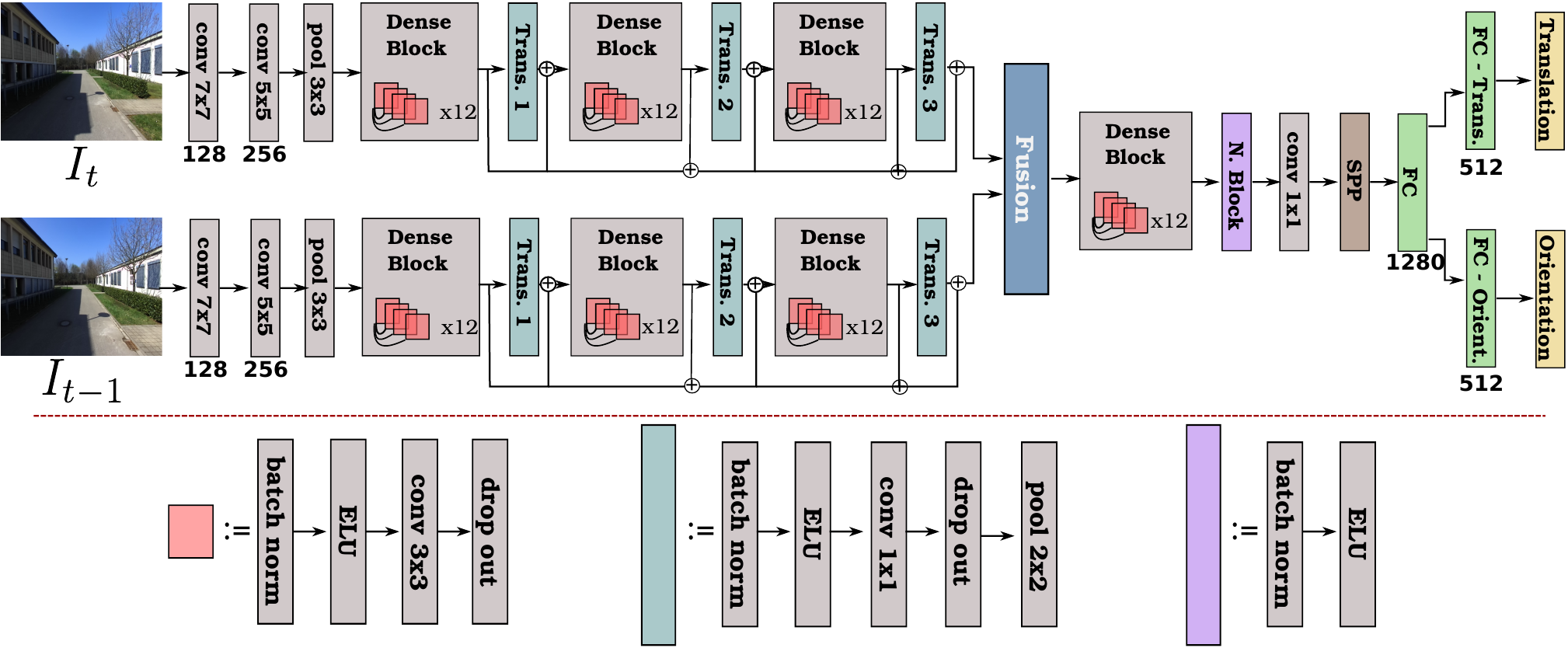}
	\caption{Depiction of the proposed VONet architecture. Each
	Dense stream is fused after three Dense Blocks. The fusion layer is 
responsible to concatenate the time dependent features into one representation 
that is fed to another Dense Block and to a SPP layer. The SPP features serve 
as input to a fully connected layer that is consequently split into specific 
layers, for translation and orientation prediction.}
	\label{fig:VONet}
\end{figure}

To estimate the visual odometry or relative camera pose we propose a Siamese 
architecture built upon dense blocks~\cite{huang2016densely} and spatial 
pyramid pooling (SPP)~\cite{SPP}. The direct connections between multiple layers of a dense block yielded state-of-the-art 
results 
on other tasks, and as we show in this paper, also yields excellent performance for the task at hand. \figref{fig:VONet} shows 
the 
proposed VONet 
architecture. The network consist of two parts: time feature representation and 
regression, respectively. The time representation streams are built upon dense 
blocks with intermediate transition blocks. Each dense block contains multiple dense
layers with direct connections from each layer to all subsequent layers. Consequently, each dense layer receives as input the 
feature maps of all preceding layers.
A dense layer is composed of four consecutive operations; namely batch normalization (Batch Norm)~\cite{ioffe2015batch},
rectified linear unit (ReLU)~\cite{nair2010rectified}, a $3\times3$ convolution (conv) and a drop-out.
The structure of the transition layer is very similar to that of a dense layer, with the addition of a $2\times2$ pooling 
operation after the drop-out and a $1\times1$ convolution instead of the $3\times3$ one.
Furthermore, we alter the first two convolutional operations in the time representation streams
by increasing their kernel sizes to $7\times7$ and $5\times5$, respectively. 
These layers serve the purpose of observing larger image areas, thus 
providing better motion prediction. We also modify 
the dense blocks by replacing ReLUs with exponential linear units (ELUs), which proved to speed up training and provided better 
results~\cite{clevert2015fast}. Both network streams are identical and learn feature representations to each of the images $I_t$ 
and $I_{t-1}$.   

We fuse both branches from VONet through concatenation. We also tried a fully connected layer, but extracting and fusing features 
from 
convolutional layers produced better results. The fused features are passed to 
another dense block, which is followed by a \emph{Spatial Pyramid Pooling} (SPP) layer. SPPs are another main building block of 
our approach. Using SPP layers has two main advantages 
for the specific task we are tackling. First, it allows the use of the presented architecture
with arbitrary image resolutions. The second advantage is the layer's ability to maintain part of the spatial information by 
pooling within local spatial bins. The final layers of our network are fully connected layers used as a regressor estimating two 
\textit{2-dimensional} vectors. 

\subsection{Topological Localization}
\label{sec:locnet}

Given a set of images acquired on a path during navigation, the task of topological localization
can be considered as the problem of identifying the location among a set of previously visited ones.
To this end, we first pre-process the acquired images to create the distinct key-frames, then we train a CNN that learns the 
probability distribution over the likelihood of the key-frames given the input image.

To extract visually distinct locations from a given path, we cluster poses by computing an
image-to-image correlation score, so that similar images are grouped together in one cluster.
We select clusters that are within a certain distance
threshold $\mathit{d_{th}}$ to represent the distinct key-frames. We chose the value of $\mathit{d_{th}}$ 
so that there is a small visual aliasing between the generated key-frames.

\begin{figure}[t]
	\centering
	\includegraphics[width=\textwidth]{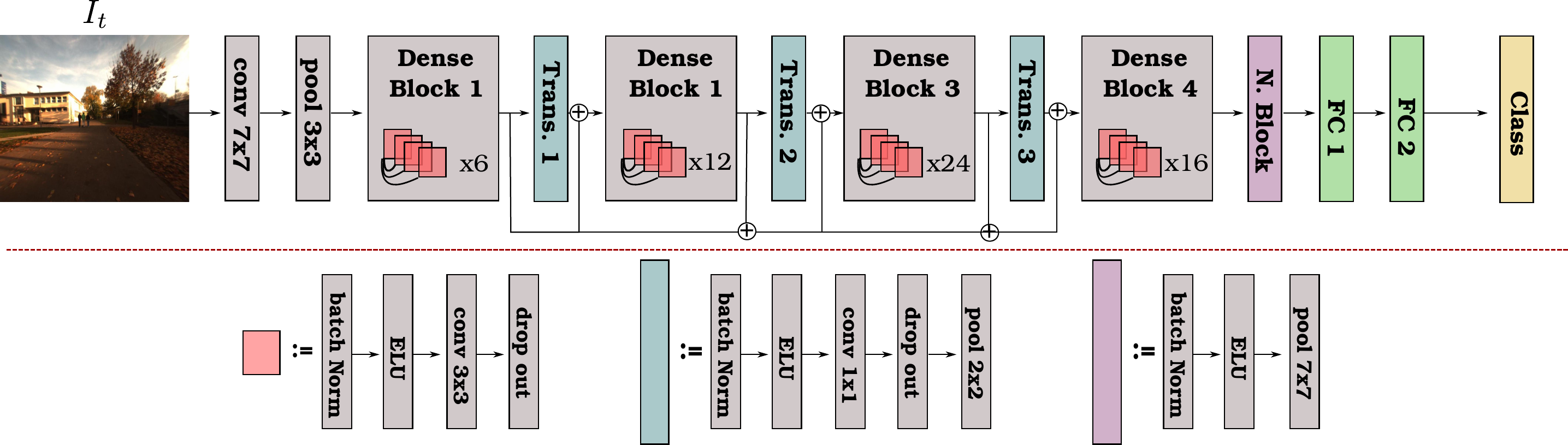}
	\caption{Proposed LocNet-121 architecture. At each Dense Block feature maps from previous
	layers are fused together as input. The lower legend shows the constituting parts for each block.}
	\label{fig:LocNetArch}
\end{figure}

%
Similar to our VONet, we introduce a network architecture based on DenseNet~\cite{huang2016densely},
namely LocNet. Our core network consists of four dense blocks with intermediate
transition blocks.  
The proposed architecture differs from the DenseNet architecture by the addition of an
extra fully connected layer before the prediction, and the addition of extra connections between the 
dense blocks fusing information from earlier layers to later ones.
Similar to the experiments of Huang \emph{et al.}~\cite{huang2016densely} on ImageNet, we experiment with the different 
depths and growth rates for the proposed architecture, using the same configurations as the ones reported by Huang \emph{et al.}.
\figref{fig:LocNetArch} illustrates the network architecture for LocNet-121. 
Given an input image, the network estimates the probability  distribution over the 
discrete set of locations.

\subsection{Topometric Localization}
\label{sec:topometricMethodology}

Our topometric approach aims to refine metric localization given topological 
priors. The topological localization network provides a set of values 
corresponding to locations and the probability of prediction confidence. For 
this purpose we need to fuse metric and topological network predictions into a 
single representation $C$. The proposed topometric fusion approach is 
optimized as below: 

\begin{equation}
  C := F + B + \lambda S
\end{equation}
where $F$ is the forward drift correction, $B$ the backward path optimization, 
 and $\lambda >0$ the smooth parameter of the $S$ smoothness term. At time 
$t\geq0$, $\mathbf{x}_t \in 
SE(2)$ is the pose and $\mathbf{n}_t \in 
SE(2)$ is the matched topological node with probability higher than a threshold 
$\delta$.     
\begin{align}
 \label{forward}
 F&:= \left \| \mathbf{x}_{t}^{tr} - \mathbf{n}_{t}^{tr} \right \|_2^2 + 
 | \mathbf{x}_{t}^{\theta} - \mathbf{n}_{t}^{\theta} |^2,
   \\
  \label{backward}  
 B&:= \sum_{t-t_w\leq\tau\leq t-1} e^{-\lambda_{tr}(t-\tau)}  \left \| 
\mathbf{x}_{\tau}^{tr} - \mathbf{n}_{\tau}^{tr} \right \|_2^2 + 
e^{-\lambda_{\theta}(t-\tau)} 
  | \mathbf{x}_{\tau}^{\theta} - \mathbf{n}_{\tau}^{\theta} |^2,
 \\
 \label{smoothness}
S&:= \argmin_{\alpha \in \mathbb{R}^3} \sum_{0\leq\tau\leq t} \left \| 
\mathbf{x}_{\tau}^{tr} - P_{\alpha}\left (\tau \right ) \right \|_2^2
W\left ( \tau \right ),
\end{align}
where $P_{\alpha}\left(\tau \right )$ is a quadratic polynom 
and 
$W\left(\tau\right )$ is the weight function described in 
\cite{locallyweightRegression}.  

\eqref{forward} presents our forward drift correction approach, which
we model in the following manner: Given a high probability topological node matching, we
compute 
the translation and rotation errors and propagate it through the next metric 
localization predictions until another topological node is 
detected. Whereas the forward drift correction component is responsible for 
mitigating future error accumulation, the backward and smoothness terms are designed to 
further correct the obtained trajectory. The backward path optimization 
is introduced in~\eqref{backward}. The backward optimization term works 
as follows: Given a confident topological node it calculates its error to 
it and, using an exponential decay, corrects previous predictions in 
terms of translation and rotation values, until it reaches a predefined time 
window $t_w$. We also treat the exponential decay functions separately for 
translation and orientation because of their different scale values.

The final term, which compromises smoothing the trajectory, is presented in 
\eqref{smoothness}. It corresponds to a 
local regression approach, similar to the that used by Cleveland~\cite{locallyweightRegression}. Using a 
quadratic polynomial model with $\alpha \in \mathbb{R}^3$  we locally fit a 
smooth surface to our current trajectory. One 
difference from this term to the others is that such term is only applied to 
translation. Rotation is not optimized in this term given the angles are 
normalized. We choose smoothing using local regression due to the flexibility 
of the technique, which does not require any specification of a function to fit 
the model, only requiring a smoothing parameter and the degree of the local 
polymonial.              

\section{Experiments}

We evaluated our topometric localization approach on a dataset collected from 
Freiburg campus across different seasons.
The dataset is split into two parts; RGB data and RGB-D data. We perform a separate evaluation for each the proposed 
Visual Odometry and Topological Localization networks, as well as the 
fused approach. The implementation was based on the publicly available 
Tensorflow learning toolbox~\cite{abadi2016tensorflow}, and all 
experiments were carried out with a system containing an NVIDIA Titan X GPU.

\subsection{Experimental setup - Dataset}
In order to evaluate the performance of the suggested approach, we introduce the Freiburg Localization 
(FLOC) Dataset, where we use our robotic platform, Obelix~\cite{kummerle2015autonomous} for the data collection.
Obelix is equipped with several sensors, however in this work, we relied on 
three laser scanners, a
Velodyne HDL-32E scanner, the Bumblebee camera and a vertically mounted
SICK scanner. Additionally, we mounted a ZED stereo camera to obtain depth data.
As previously mentioned, the dataset was split into two parts; RGB and RGB-D data. We used
images from the Bumblebee camera to collect the former, and the ZED camera for the latter. 
The dataset collection procedure went as follows; we navigate the robot along a chosen path twice,
starting and ending at the same location. One run is chosen for training and the other for
testing. The procedure is repeated several times for different paths on campus, at different
times of the day throughout a period of six months. The collected dataset has a 
high degree of noise incurred by pedestrians and cyclists walking by in different directions
rendering it more challenging to estimate the relative change in motion between frames.
We use the output of the SLAM system of Obelix
as a source of ground-truth information for the traversed trajectory. We select nodes that are 
at a minimum distance of $1~m$ away from each other, along with the corresponding camera frame. Each node
provides the $3\mathit{D}$ position of the robot, and the rotation in the form of a quaternion. 
As mentioned previously in~\secref{sec:metricMeth}, we opt for representing the rotations in Euler6 notation.
Accordingly, we convert the poses obtained from the SLAM output of Obelix to Euler6. Furthermore,
we disregard translation motion along the $\mathit{z}$-axis, and rotations along the $\mathit{x}$- and $\mathit{y}$-
axes, as they are very unlikely in our setup.

For the remainder of this section, we focus our attention on two sequences of the FLOC dataset,
namely Seq-1 and Seq-2. Seq-1 is an RGB-D sequence with 
$262$ meters of total length captured by the ZED camera, while Seq-2 
is comprised of a longer trajectory of $446$ meters of RGB only data captured by the 
Bumblebee camera. We favored those two sequences from the dataset as they are representative 
of the challenges faced by vision-based localization approaches. Seq-1 represents a case
where most vision-based localization systems are likely to perform well as the trajectory
length is short. Moreover the presence of depth information facilitates the translation estimation
from the input images. On the other hand, Seq-2 is more challenging with almost double the trajectory
length and no depth information.

\subsection{Network Training}

The networks were training on a single stage manner. VONet was trained using 
Adam solver~\cite{AdamSolver}, with a mini-batch of $2$ for $100$ epochs. The 
initial learning rate is set to $0.001$, and is multiplied by $0.99$ every two 
epochs. The input resolution image is downscale to $512\times384$ due to memory 
limitations. We adopt the same weight initilization as in 
\cite{huang2016densely}. The loss function balance variable $\beta$ is set to 
$10$. The training time for VONet for 100 epochs took around $12$ hours on a single 
GPU. 
LocNet was trained using Nesterov Momentum
~\cite{sutskever2013importance}, with a base learning rate of $0.01$ for $10$
epochs with a batch size of $10$. The learning rate was fixed throughout  
the evaluation.

 \subsection{Metric Localization}
 
We evaluate our VO based metric localization approach over multiple sequences 
of FLOC. For these experiments we compared 
our approach with Nicolai  \emph{et al.}~\cite{NicolaiRSSw2016}, 
DeepVO~\cite{MohantyDeepVO} and cnnBspp~\cite{melekhov2017cnnBspp}. For each 
sequence, two metrics are provided: average translation error and average 
rotation error as a function of the sequence length.      

\begin{figure}[t]
	\centering
	
\includegraphics[width=0.75\textwidth]{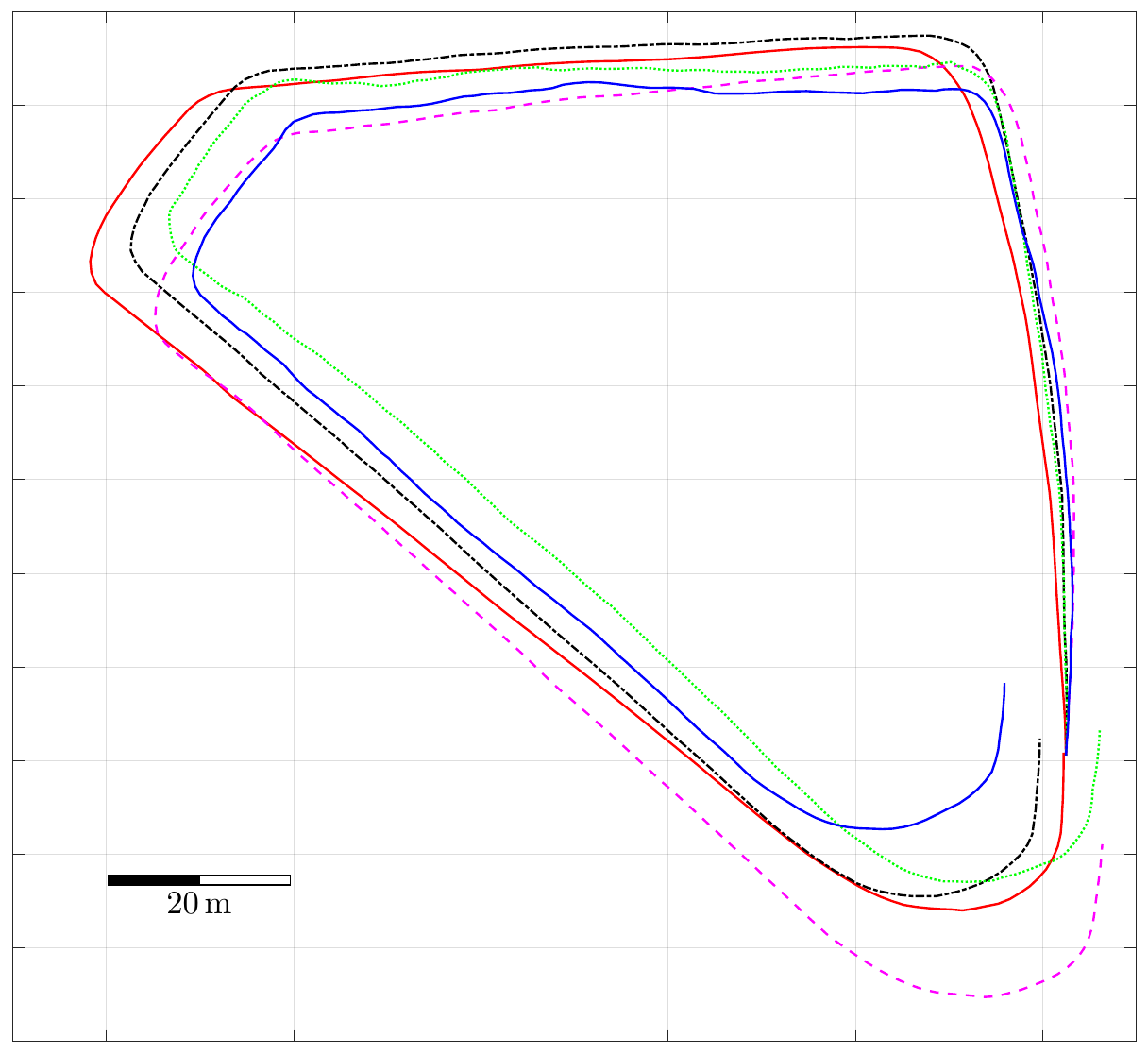}
	\boxplotlegend
	\caption{Predicted trajectories vs Ground Truth for Seq-1.}
	\label{fig:shortpath}
\end{figure}

\figref{fig:shortpath} shows the computed trajectories of the compared 
methods for Seq-1. \tabref{tab:shortpath} depicts the average 
translation and rotation error as a function of sequence $1$ length. Our 
approach outperforms the compared methods with almost two times 
smaller error for translation and rotation
inference making it the closest to the ground-truth.


\begin{table}
\centering
\caption{Average translation and rotation as a function of sequence length 
(Seq-1).}
\label{tab:shortpath}
\begin{tabular}{p{3cm}p{2cm}p{2cm}}
\hline\noalign{\smallskip}
Method & Translation [\%] &Rot [deg/m]\\
\noalign{\smallskip}\hline\hline\noalign{\smallskip}
Nicolai~\cite{NicolaiRSSw2016} & 3.12 & 0.6518 \\
DeepVO~\cite{MohantyDeepVO} & 2.41  & 0.4471 \\
cnnBspp~\cite{melekhov2017cnnBspp} & 3.09 & 0.4171 \\
Ours & \bf{1.54} & \bf{0.2919} \\
\noalign{\smallskip}\hline\noalign{\smallskip}
\end{tabular}
\end{table}

\begin{figure}[t]
	\centering
	\includegraphics[width=0.75\textwidth]{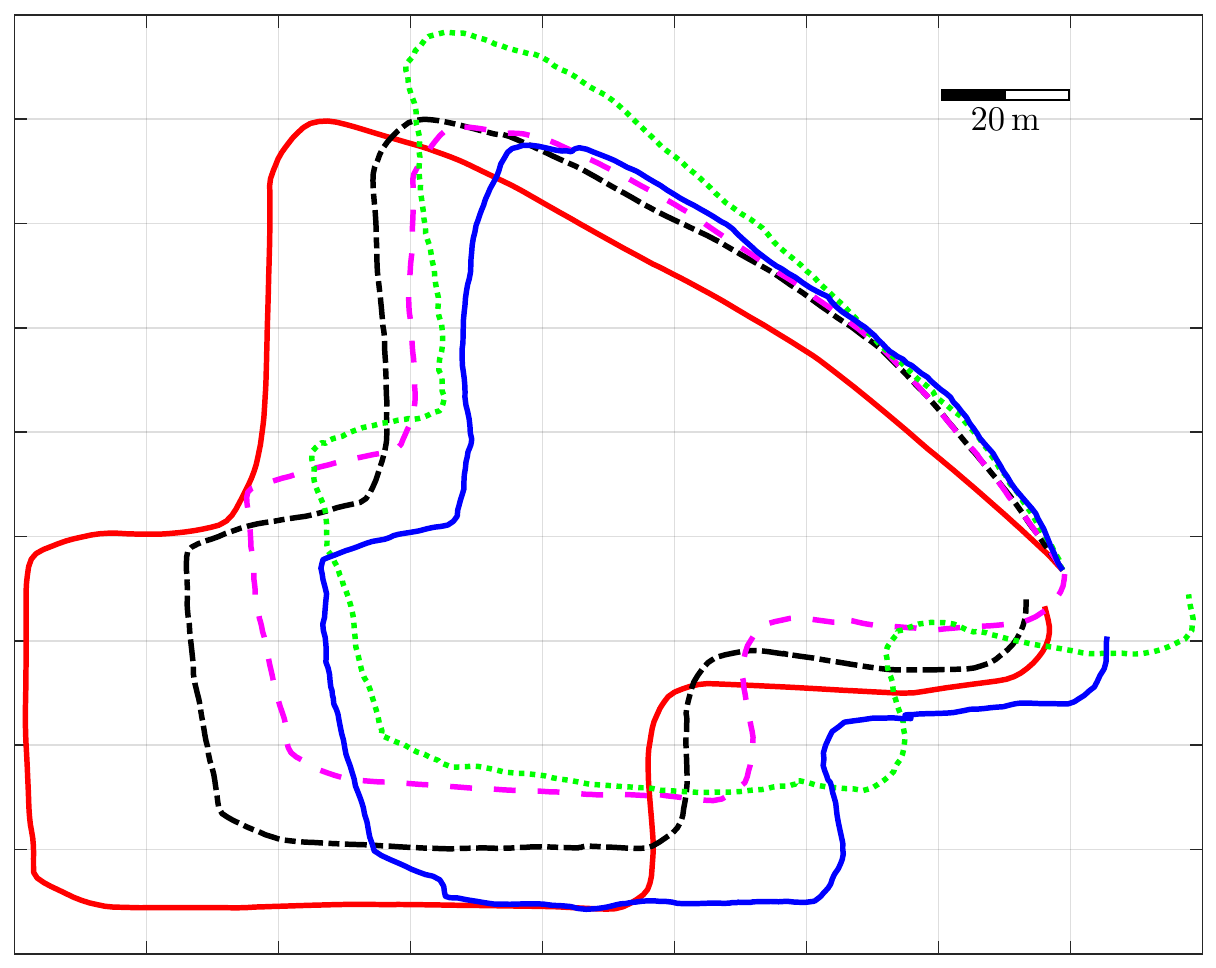}
	\boxplotlegend
	\caption{Predicted trajectories vs Ground Truth for Seq-2.}
	\label{fig:longpath}
\end{figure}

Seq-1 shows that our VO approach can achieve state-of-the-art 
performance. However the cumulative error characteristic of the problem
makes it harder for longer trajectories. \figref{fig:longpath} presents 
results for Seq-2.
For this experiment 
the trajectories have a bigger error, 
especially for translation. \tabref{tab:longpath} quantifies the obtained values,
confirming the difficulty of 
this sequence for all tested techniques. The results show that our approach is 
still capable of largely outperforming the compared methods, with a translation 
and rotation error almost twice as low as the other methods. 
Despite the performance of our approach, it is still far from being competitive with
LiDAR based approaches, like 
the one used to generate our ground-truth \cite{kummerle2015autonomous}. With this goal
in mind, we exploit the topological localization method to refine our metric 
approach providing an even more precise topometric approach.


\begin{table}
\centering
\caption{Average translation and rotation as a function of sequence length 
(Seq-2).}
\label{tab:longpath}
\begin{tabular}{p{3cm}p{2cm}p{2cm}}
\hline\noalign{\smallskip}
Method & Translation [\%] &Rot [deg/m]\\
\noalign{\smallskip}\hline\hline\noalign{\smallskip}
Nicolai~\cite{NicolaiRSSw2016} & 6.1 & 0.4407 \\
DeepVO~\cite{MohantyDeepVO} & 9.13  & 0.2701 \\
cnnBspp~\cite{melekhov2017cnnBspp} & 6.11 & 0.2083 \\
Ours & \bf{3.82} & \bf{0.1137} \\
\noalign{\smallskip}\hline\noalign{\smallskip}
\end{tabular}
\end{table}

\subsection{Topological Localization}

In this section, we evaluate the performance of the proposed LocNet architecture. To get
an estimate of the suitability of the proposed architecture to the problem at hand,
we used the Places2 dataset~\cite{zhou2016places} for scene recognition. The dataset contains
over ten million scene images divided into 365 classes. We use the pretrained DenseNet model
on ImageNet to initialize the weights for our LocNet-121 architecture, as our architecture
is quite similar to DenseNet aside from using a different activation function. Using the
pretrained model, we are able to achieve comparable performance to the Places365-GoogLeNet
architecture as reported by the authors.
Additionally, we compare the performance of our LocNet architecture with that of Residual
Networks (ResNet)~\cite{he2016deep} given its recent performance in image recognition.
We evaluate the performance of both architectures over multiple sequences of the FLOC dataset
and the Cambridge Landmarks dataset~\cite{kendall2015posenet}. For both datasets, we report
the accuracy in terms of the number of images where the predicted location is within a 
$1~m$ radius of the ground-truth pose. 
\tabref{tab:performanceCampus} illustrates the performance results on Seq-1 of the FLOC
dataset. We investigate the effect of the depth of the network on the accuracy of the predicted poses, while
comparing the number of parameters. The best performance was achieved using LocNet-169
with an accuracy of $90.4\%$ with approximately $3\times$ less parameters than its
best performing counterpart in ResNet.
\tabref{tab:performancePoseNet} illustrates the performance on the different scenes from the 
Cambridge Landmarks dataset. On this dataset, LocNet-201 achieves the best performance
with the exception of King's College scene. It is worth noting that LocNet-169 achieves
the second highest accuracy in four out of the five remaining scenes, providing further
evidence to the suitability of this architecture to the problem at hand.
For the remainder of the experimental evaluation, we use the prediction output from LocNet-201.

\begin{table}
\centering
\caption{Classification accuracy of the different networks on the Freiburg Campus Dataset (Seq-1).}
\label{tab:performanceCampus}
\begin{tabular}{p{3cm}p{2cm}p{2cm}p{2cm}}
\hline\noalign{\smallskip}
Classifier & Depth & Params & Accuracy\\
\noalign{\smallskip}\hline\hline\noalign{\smallskip}
ResNet				 & 34 & 22M & 84.3\% \\
ResNet				 & 50 & 26M & 84.7\% \\
ResNet				 & 101 & 44M & 85.1\% \\
ResNet				 & 152 & 60M & 85.1\% \\
LocNet               & 121 & 9M & 90.0\% \\
LocNet               & 169 & 15M & \textbf{90.4\%} \\
LocNet               & 201 & 20M & 88.99\% \\
\noalign{\smallskip}\hline\noalign{\smallskip}
\end{tabular}
\end{table}

\begin{table}
\centering
\caption{Classification accuracy of the different networks on the PoseNet Dataset.}
\label{tab:performancePoseNet}
\begin{tabular}{p{2.3cm}p{1.2cm}p{1.2cm}p{1.2cm}p{1.2cm}p{1.2cm}p{1.2cm}p{1.2cm}}
\hline\noalign{\smallskip}
Scene            &Classes & RN-34 & RN-50 & RN-101 & LN-121 & LN-169 & LN-201\\
\noalign{\smallskip}\hline\hline\noalign{\smallskip}
Shop Facade      & 9    		    & 83.8\%    & 82.8\%    & 77.8\%     & 80.3\%	   & 85.7\%       & 
\textbf{86.6\%}\\
King's College   & 20            & 86.7\%    & 87.7\%    & 85.1\%	    & 87.1\%       & \textbf{90.4\%}& 89.5\%\\
Great Court      & 109            & 69.9\%    & 67.4\%    & 65.6\%     & 68.0\%       & 68.8\%        & \textbf{71.44\%} \\
Old Hospital     & 14            & 81.4\%    & 86.0\%    & 84.0\%     & 88.6\%       & 90.1\%       & \textbf{90.6\%}\\
St.Mary's Church & 30            & 78.4\%    & 79.0\%    & 79.1\%     & 87.7\%       & 88.4\%        & \textbf{88.8\%} \\
Street           & 241           & 68.4\%    & 73.4\%    & 72.7\%     & 67.1\%       & 73.3\%        & \textbf{75.6\%}\\
\noalign{\smallskip}\hline\noalign{\smallskip}
& & & & & &\multicolumn{2}{c}{\scriptsize{RN: ResNet, LN: LocNet}}
\end{tabular}
\end{table}

\subsection{Topometric Localization}

\begin{figure}[t]
	\centering
	
\includegraphics[width=0.75\textwidth]{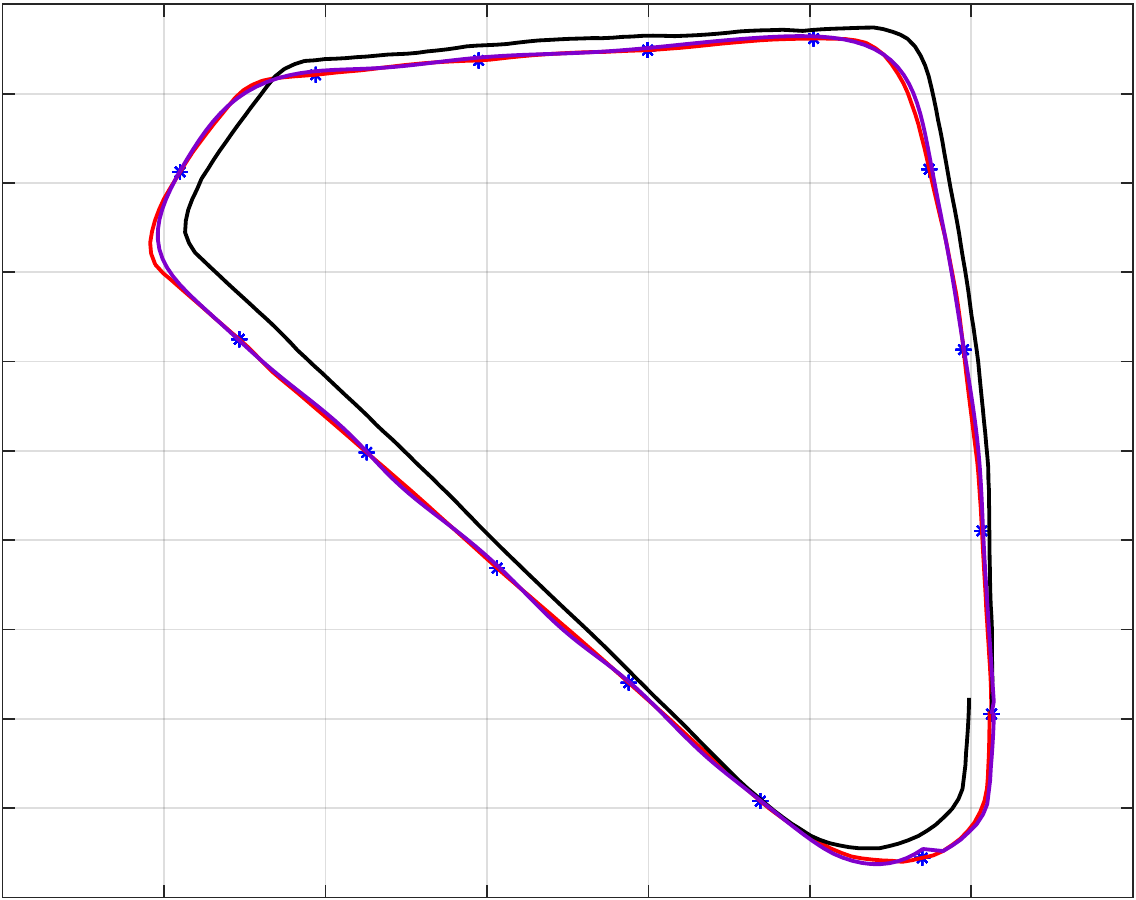}
	\boxplotlegendTopoMetric
	\caption{Seq-1, metric vs topometric.}
	\label{fig:shortpathTopometric}
\end{figure}

This section presents the results of fusing both topological and metric localization techniques. 
\figref{fig:shortpathTopometric} presents both the 
metric and topometric results for Seq-1. As can be noticed the 
trajectory difference between ground truth and our topometric approach is 
almost not visually distinguishable. \tabref{tab:topometric} shows an 
improvement of $7\times$ in the translation inference and superior to $6\times$
for orientation. Such values provide competitive results even to the 
LiDAR system utilized to provide ground-truth to FLOC.

\begin{figure}[t]
	\centering
	
\includegraphics[width=0.75\textwidth]{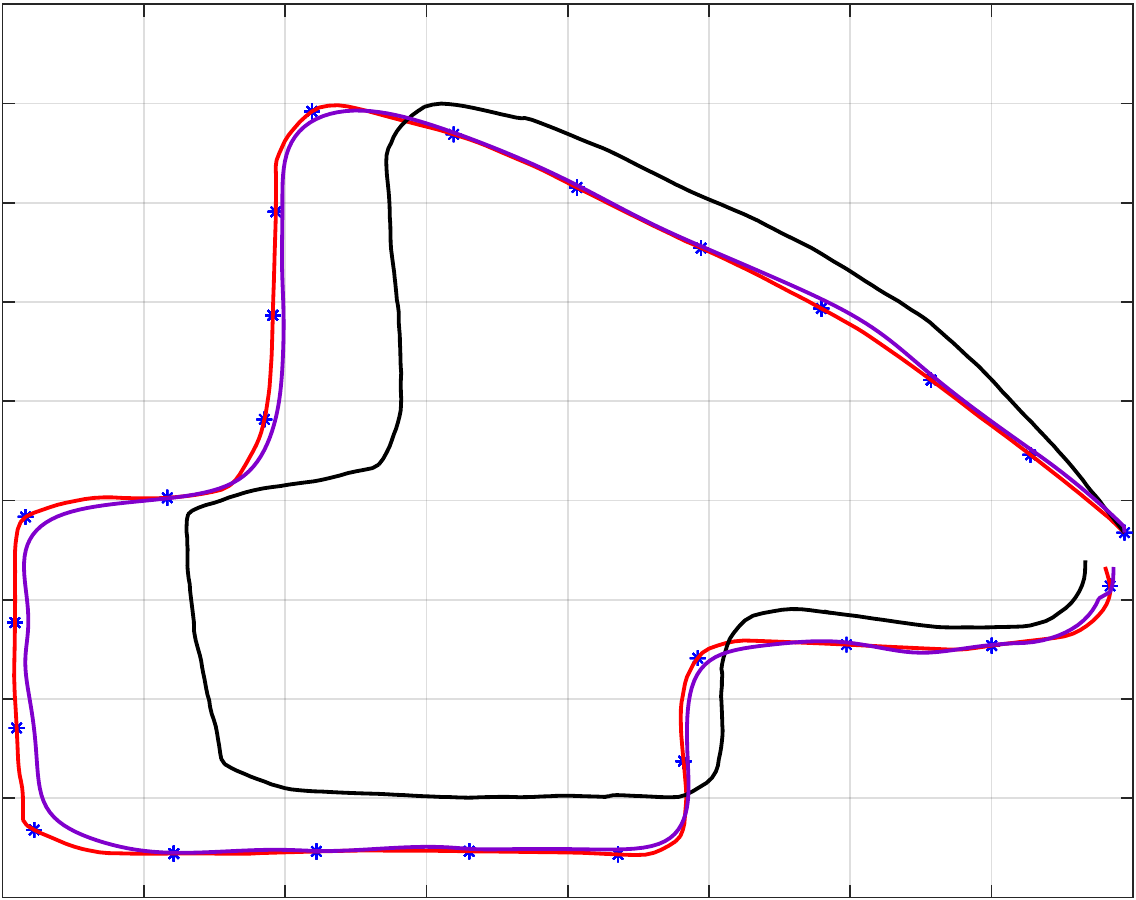}
	\boxplotlegendTopoMetric
	\caption{Seq-2, metric vs topometric.}
	\label{fig:longpathTopometric}
\end{figure}

We also evaluated topometric localization using Seq-2. 
\figref{fig:longpathTopometric} depicts the obtained results. While the
results for this sequence are not as accurate as those of Seq-1, the gain 
in translation is more than $10\times$ the metric counterpart. For 
orientation, even though our metric approach already presents good results, the error is 
reduced by half using our topometric technique, as shown 
in~\tabref{tab:topometric}.

\begin{table}
\centering
\caption{Metric vs Topometric Results.}
\label{tab:topometric}
\begin{tabular}{p{3cm}p{2cm}p{2cm}}
\hline\noalign{\smallskip}
Method & Translation [\%] &Rot [deg/m]\\
\noalign{\smallskip}\hline\hline\noalign{\smallskip}
Seq. 1 Metric  & 1.54 & 0.2919 \\
Seq. 1 Topometric & \textbf{0.21}  & \textbf{0.0464} \\ \hline
Seq. 2 Metric & 3.82 & 0.1137 \\
Seq. 2 Topometric & \textbf{0.38} & \textbf{0.0634} \\
\end{tabular}
\end{table}

One important characteristic of our topometric approach is that the error  
is bounded in between consecutive key-frames and does not grow unboundedly
over time like with the metric localization method. The presented results show that based on the 
frequency of the topological nodes we can expect a maximum cumulative error 
based on the corrected topometric error and not on the pure metric cumulative 
error.

\section{Conclusions}

In this paper, we have presented a novel deep learning based topometric localization
approach. We have proposed a new Siamese architecture, which we refer to as VONet,
to regress the translational and rotational relative motion between two 
consecutive camera images. The output of the proposed network provides the visual odometry
information along the traversed path. Additionally, we have discretized the trajectory
into a finite set of locations and have trained a convolutional neural
network architecture, denoted as LocNet, to learn the probability distribution over
the locations. We have proposed a topometric optimization technique that corrects the 
drift accumulated in the visual odometry and further corrects the traversed path. We evaluated our approach on the new Freiburg 
Localization (FLOC) dataset, which we collected over the course of six months in adverse weather conditions using different 
modalities and which we will provide to the research community. The extensive experimental evaluation shows that our proposed 
VONet and LocNet architectures surpass current state-of-the-art methods for their respective problem domain. Furthermore, using 
the proposed topometric approach we improve the localization accuracy by one order of magnitude.

\begin{acknowledgement}
This work has been partially supported by the European Commission under the 
grant numbers H2020-645403-ROBDREAM, ERC-StG-PE7-279401-VideoLearn,
the Freiburg Graduate
School of Robotics.
\end{acknowledgement}

\bibliographystyle{spmpsci}
\bibliography{references}

\end{document}